\documentclass[10pt,twocolumn,letterpaper]{article}

\usepackage{iccv}
\usepackage{times}
\usepackage{epsfig}
\usepackage{graphicx}
\usepackage{amsmath}
\usepackage{amssymb}
\usepackage{array}
\usepackage{bm}
\usepackage{booktabs}
\usepackage{caption3}
\usepackage{url}
\usepackage{multirow}
\usepackage{xcolor}
\usepackage{algorithm}
\usepackage{algorithmic}
\usepackage[caption=false,font=scriptsize]{subfig}

\usepackage[breaklinks=true,bookmarks=false]{hyperref}

\iccvfinalcopy 

\def\iccvPaperID{****} 

\ificcvfinal\fi
\begin{document}

\title{Joint Group Feature Selection and Discriminative Filter Learning for Robust Visual Object Tracking}

\author{Tianyang Xu$^{1,2}$  ~~Zhen-Hua Feng$^2$  ~~Xiao-Jun Wu$^{1*}$ ~~Josef Kittler$^2$\\
$^1$School of Internet of Things Engineering, Jiangnan University, Wuxi, China\\
$^2$ Centre for Vision, Speech and Signal Processing (CVSSP), University of Surrey, Guildford, UK\\
{\tt\small tianyang\_xu@163.com, z.feng@surrey.ac.uk, wu\_xiaojun@jiangnan.edu.cn, j.kittler@surrey.ac.uk}
}

\maketitle
\ificcvfinal\thispagestyle{empty}\fi

\begin{abstract}
We propose a new Group Feature Selection method for Discriminative Correlation Filters (GFS-DCF) based visual object tracking. The key innovation of the proposed method is to perform group feature selection across both channel and spatial dimensions, thus to pinpoint the structural relevance of multi-channel features to the filtering system. In contrast to the widely used spatial regularisation or feature selection methods, to the best of our knowledge, this is the first time that channel selection has been advocated for DCF-based tracking. We demonstrate that our GFS-DCF method is able to significantly improve the performance of a DCF tracker equipped with deep neural network features. In addition, our GFS-DCF enables joint feature selection and filter learning, achieving enhanced discrimination and interpretability of the learned filters. 

To further improve the performance, we adaptively integrate historical information by constraining filters to be smooth across temporal frames, using an efficient low-rank approximation. By design, specific temporal-spatial-channel configurations are dynamically learned in the tracking process, highlighting the relevant features, and alleviating the performance degrading impact of less discriminative representations and reducing information redundancy. The experimental results obtained on OTB2013, OTB2015, VOT2017, VOT2018 and TrackingNet demonstrate the merits of our GFS-DCF and its superiority over the state-of-the-art trackers. The code is publicly available at \url{https://github.com/XU-TIANYANG/GFS-DCF}.
\end{abstract}

\section{Introduction}
To consistently and accurately track an arbitrary object in video sequences is a very challenging task.
The difficulties are posed by a wide spectrum of appearance variations of an object in unconstrained scenarios.
Among existing tracking algorithms, Discriminative Correlation Filters (DCF-) based trackers~\cite{Henriques2015High} have exhibited promising results in recent benchmarks~\cite{Wu2013Online,Wu2015Object,mueller2016benchmark,Liang2015Encoding} and competitions such as the Visual Object Tracking (VOT) challenges~\cite{Kristan2015The,Kristan2016The,Kristan2017a,Kristan2018a}. 
\begin{figure}[t]
\begin{center}
\includegraphics[trim={0mm 12mm 62mm 0mm},clip,width=1\linewidth]{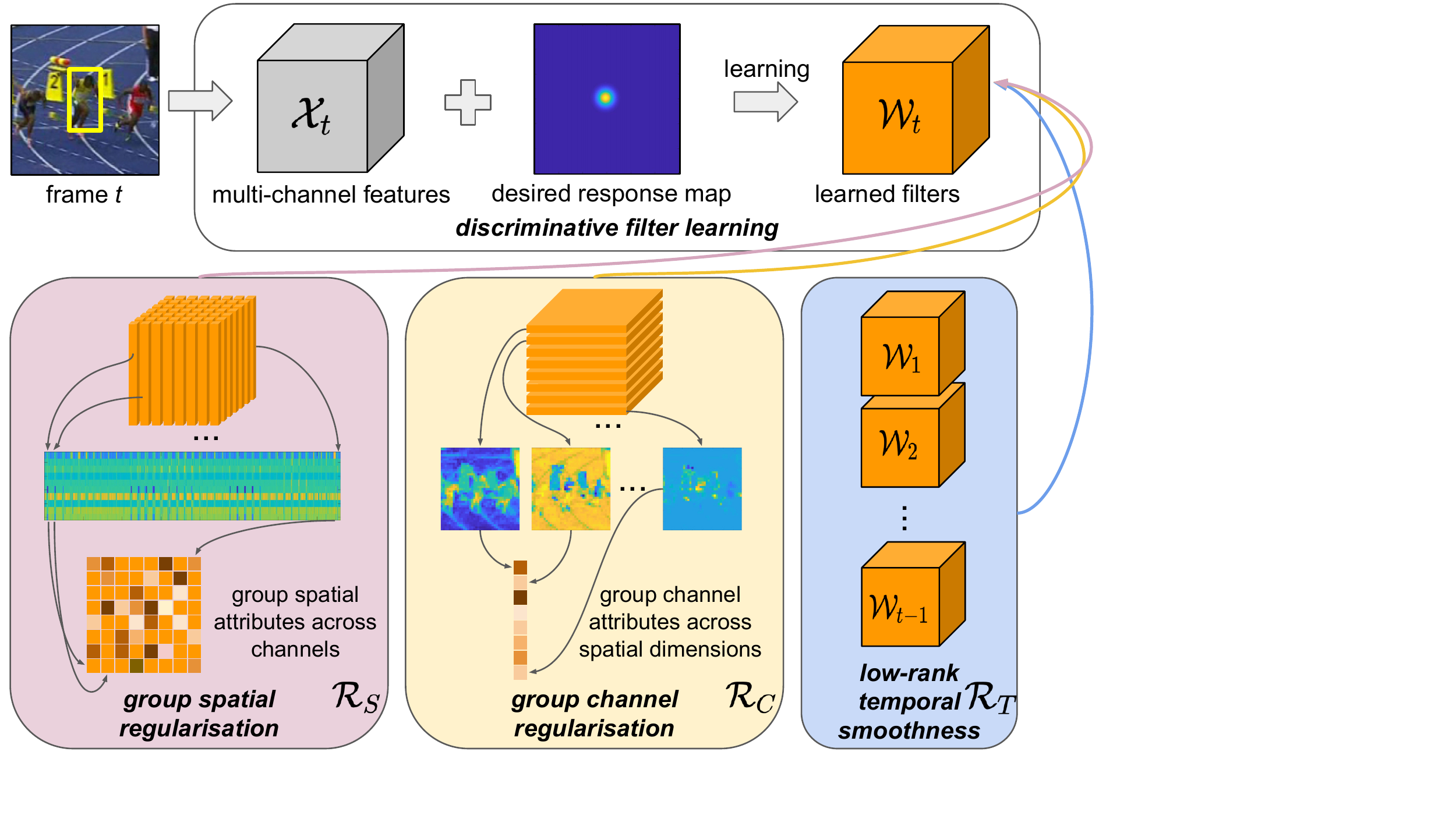}
\end{center}
\caption{In contrast to the classical DCF paradigm, our GFS-DCF performs channel and spatial group feature selection for the learning of correlation filters. Group sparsity is enforced in the channel and spatial dimensions to highlight relevant features with enhanced discrimination and interpretability. Additionally, a low-rank temporal smoothness constraint is employed across temporal frames to improve the stability of the learned filters.}
\label{framework}
\end{figure}

The success of high-performance DCF trackers is attributed to three aspects: spatial regularisation, temporal smoothness and robust image feature representation. 
Regarding the first point, as natural images and videos are projections from a 3D space into a 2D plane, spatial regularisation directly improves the tracking accuracy by potentially endowing the learned filters with a specific attention mechanism, enhancing the discrimination by focusing on less ambiguous regions~\cite{Danelljan2015Learning,Lukezic2017Discriminative,Galoogahi2017Learning,zhang2018visual}. 
Second, based on the fact that video sequences are formed by discrete image sampling of continuous dynamic scenes, reflecting the temporal smoothness of successive frames in the construction of appearance models has been shown to improve their generalisation capacity~\cite{danelljan2016adaptive,li2018learning,Danelljan2016ECO,Danelljan2016Beyond}. 
Third, with the development of robust image feature extraction methods, \textit{e.g.} Histogram of Oriented Gradient (HOG)~\cite{dalal2005histograms}, Colour Names (CN)~\cite{Weijer2009Learning} and Convolutional Neural Network (CNN) features~\cite{krizhevsky2012imagenet,szegedy2015going,lu2018deep,dong2018hyperparameter}, the performance of DCF-based trackers has been greatly improved~\cite{bhat2018unveiling,Kristan2018a,Kristan2017a}.
It is indisputable that recent advances in DCF-based tracking owe to a great extent to the use of robust deep CNN features.

Despite the rapid progress in visual tracking by equipping the trackers with robust image features, the structural relevance of multi-channel features to the filtering system has not been adequately investigated.
In particular, due to the limited number of training samples available for visual tracking, DCF-based trackers usually use a deep network pre-trained on other computer vision tasks, such as VGG~\cite{simonyan2014very} or ResNet~\cite{he2016deep} trained on ImageNet~\cite{russakovsky2015imagenet}.
In such a case, the extracted deep feature channels (maps) for an arbitrary object, which may exceed thousands, may not be compact. They may include irrelevant as well as redundant descriptors and their presence may degrade the target detection performance.
However, the tension between discrimination, information relevance, information redundancy and high-dimensional feature representations has not been systematically studied in the existing DCF paradigm.
We argue it is absolutely crucial to perform dimensionality reduction along the channel dimension to suppress irrelevant features as well as redundancy for deep neural network features.

To redress the above oversight, we propose a new Group Feature Selection method for DCF-based visual object tracking, namely GFS-DCF.
To be more specific, we reduce the information redundancy and irrelevance of high-dimensional multi-channel features by performing group feature selection across both spatial and channel dimensions, resulting in compact target representations.
It should be highlighted that our GFS-DCF differs significantly from existing DCF-based trackers, in which only spatial regularisation or selection is used.
Additionally, as supervised frame-to-frame data fitting may create excessive variability in prediction, we constrain a learned predictor (filter) to be smooth across the time dimension (frames).

Fig.~\ref{framework} depicts the basic learning scheme of the proposed GFS-DCF method.
Given the predicted location of an object in the $t$th frame, we first extract multi-channel features.
Then the extracted features and desired response map are used to learn the correlation filters for the prediction of the target in the next frame. In the filter learning stage, the combination of channel-spatial group feature selection and low-rank constraints adaptively identifies a specific temporal-spatial-channel configuration for robust discriminative filter learning.
As a result, relevant features are highlighted to improve discrimination and decrease redundancy. The main contributions of our GFS-DCF method include:
\begin{itemize}
\item A new group feature selection method for multi-channel image representations, reducing the dimensionality across both spatial and channel dimensions. To the best of our knowledge, this is the first work that considers feature compression along both spatial and channel dimensions.
According to our experiments, the proposed group channel feature selection method improves the performance of a DCF-based tracker significantly when using deep CNN features.
\item A temporal smoothness regularisation term used to obtain highly correlated filters among successive frames. To this end, we use an efficient low-rank approximation, forcing the learned filters to lie in a low-dimensional manifold with consistent temporal-spatial-channel configurations. 
\item A comprehensive evaluation of GFS-DCF on a number of well-known benchmarks, including OTB13/15~\cite{Wu2013Online,Wu2015Object}, VOT17/18~\cite{Kristan2017a,Kristan2018a}, and TrackingNet~\cite{muller2018trackingnet}.
The results demonstrate the merits of GFS-DCF, as well as its superiority over the state-of-the-art trackers. 
\end{itemize}

 
\section{Related Work}
Existing visual object tracking approaches include template matching~\cite{Lucas1981An}, statistical learning~\cite{avidan2004support}, particle filters~\cite{arulampalam2002tutorial}, subspace learning~\cite{ross2008incremental}, discriminative correlation filters~\cite{Henriques2015High}, deep neural networks~\cite{qi2016hedged} and Siamese networks~\cite{zhu2018distractor,wang2018learning,li2018high}. 
In this section, we focus on DCF-based approaches due to their outstanding performance as evidenced by recent tracking competitions such as VOT~\cite{Kristan2016The, Kristan2018a}.
For the other visual tracking approaches, readers are referred to comprehensive reviews~\cite{Smeulders2014Visual,Wu2015Object,Kristan2016The,Li2016NUS} .

One of the seminal works in the development of DCF is MOSSE~\cite{Bolme2010Visual}, which formulates the tracking task as discriminative filter learning~\cite{bolme2009average} rather than template matching~\cite{briechle2001template}.
The concept of circulant matrix~\cite{Gray2006Toeplitz} is introduced to DCF by CSK~\cite{Henriques2012Exploiting} with a padded search window, which generates more background samples for the learning stage.
Additionally, spatial-temporal context~\cite{Zhang2014Fast} and kernel tricks~\cite{Henriques2015High} are used to improve the learning formulation with the consideration of local appearance and nonlinear metric, respectively.
The DCF paradigm has further been extended by exploiting scale detection~\cite{li2014scale,danelljan2014accurate,danelljan2017discriminative}, structural patch analysis~\cite{li2015reliable,Liu2015Real,Liu2016Structural}, multi-clue fusion~\cite{tang2015multi,Ma2015Long,Hong2015MUlti,Bertinetto2016Staple,tang2018high}, sparse representation~\cite{Zhang2016In,zhang2018robust}, support vector machine~\cite{wang2017large,zuo2018learning}, enhanced sampling mechanisms~\cite{zhang2017multi,mueller2017context} and end-to-end deep neural networks~\cite{valmadre2017end,song-iccv17-CREST}. 

Despite the great success of DCF in visual object tracking, it is still a very challenging task to achieve high-performance tracking for an arbitrary object in unconstrained scenarios.
The main obstacles include: spatial boundary effect, limited feature representation capacity and temporal filter degeneration.

To alleviate the boundary effect problem caused by the circulant structure, SRDCF~\cite{Danelljan2015Learning} stimulates the interest in spatial regularisation~\cite{danelljan2016adaptive,Danelljan2016Beyond,Danelljan2016ECO,li2018learning}, which allocates more energy for the central region of a filter using a predefined spatial weighting function. 
A similar idea has been pursued by means of pruning the training samples or learned filters with a predefined mask~\cite{Galoogahi2015Correlation,Lukezic2017Discriminative,Galoogahi2017Learning,li2018learning}.
Different from those approaches, to achieve spatial regularisation, LSART forces the output to focus on a specific region of a target~\cite{sun2018learning}.
A common characterisation of the above approaches is that they are all based on a fixed spatial regularisation pattern, for example, a predefined mask or weighting function.
To achieve adaptive spatial regularisation, LADCF~\cite{xu2018learning} embeds dynamic spatial feature selection in the filter learning stage. Thanks to this innovation it has achieved the best results in the public VOT2018 dataset~\cite{Kristan2018a}.
The above spatial regularisation methods decrease the ambiguity emanating from the background and enable a relatively large search window for tracking.
Nevertheless, these approaches only consider information compression along the spatial dimension.
In contrast, our GFS-DCF method performs group feature selection along both the channel and spatial dimensions, resulting in more compact object appearance descriptions.

Second, as feature representation is the most essential factor from the point of view of high-performance visual tracking~\cite{wang2015understanding}, combinations of hand-crafted and deep features have widely been used in DCF-based trackers~\cite{Danelljan2014Adaptive,Henriques2015High,bhat2018unveiling}. However, the structural relevance of multi-channel features in the filter learning system has not been considered. 
The redundancy and interference from the high-dimensional representations impede the effectiveness of learning dense filters.
To unify the process of information selection across the spatial and channel dimensions, our GFS-DCF performs group feature selection and discriminative filter learning jointly. 

Last, to mitigate temporal filter degeneration, historical clues are reflected in SRDCFdecon~\cite{danelljan2016adaptive} and C-COT~\cite{Danelljan2016Beyond}, with enhanced robustness and temporal smoothness, by gathering multiple previous frames in the filter learning stage. 
To alleviate the computational burden, ECO~\cite{Danelljan2016ECO} manages the inherent computational complexity by clustering historical frames and employing projection matrix for multi-channel features. 
Our GFS-DCF, on the other hand, is robust to temporal appearance variations by constraining the learned filters to be smooth across frames using an efficient low-rank approximation.
Consequently, the relevant spatial-channel features are consistently highlighted in a dynamic low-dimensional subspace. 

\section{DCF-based Visual Object Tracking}
Given the initial location of an object in a video, the aim of visual object tracking is to localise the object in the successive video frames.
Assume we have the estimated location of the object in the $t$th frame. To localise the object in the $t+1$th frame, DCF~\cite{Henriques2015High} learns multi-channel filters, $\mathcal{W}_t\in\mathbb{R}^{N\times N\times C}$, using a pair of training samples $\{\mathcal{X}_t,\mathbf{Y}\}$, where $\mathcal{X}_t\in\mathbb{R}^{N\times N\times C}$ is a tensor consisting of $C$-channel features extracted from the $t$th frame and $\mathbf{Y}\in\mathbb{R}^{N\times N}$ is the desired response map identifying the object location. 
To obtain $\mathcal{W}_t$, DCF formulates the objective as a regularised least square problem:
\begin{equation}
\widetilde{\mathcal{W}_t} = \arg\underset{\mathcal{W}_t}{\min}
\left\|\sum\limits_{k=1}^C\mathbf{W}^k_t\circledast{\mathbf{X}^k_t}-\mathbf{Y}\right\|_F^2 + \mathcal{R}\left(\mathcal{W}_t\right),
\label{obj}
\end{equation}
where $\circledast$ is the circular convolution operator~\cite{Henriques2012Exploiting},
$\mathbf{X}^k_t \in\mathbb{R}^{N\times N}$ is the $k$-th channel feature representation, $\mathbf{W}^k_t \in\mathbb{R}^{N\times N}$ is the corresponding discriminative filter and $\mathcal{R}\left(\mathcal{W}_t\right)= \lambda\sum_{k=1}^C\|\mathbf{W}^k_t\|^2_F$ is a regularisation term.
A closed-form solution to the above optimisation task can efficiently be obtained in the frequency domain~\cite{Henriques2015High}.

In the tracking stage, the filters learned from the first frame are directly used to localise the object in the second frame.
For the other frames, the filers are updated as:
\begin{equation}\label{up}
\mathcal{W}_t = \alpha\widetilde{\mathcal{W}_t}+\left(1-\alpha\right)\mathcal{W}_{t-1},
\end{equation}
where $\alpha \in [0,1]$ is a pre-defined updating rate.
Given a search window in the $(t+1)$st frame, we first extract multi-channel features, $\mathcal{X}_{t+1}$. 
Then the learned correlation filters, $\mathcal{W}_t$, from the $t$th frame are used to estimate the response map in the frequency domain efficiently:
\begin{equation}
\hat{\textbf{R}}=\sum\limits_{k=1}^{C}\hat{\mathbf{X}}^k_{t+1}\odot\hat{\mathbf{W}}^k_{t},
\end{equation}
where $\hat{\cdot}$ denotes Discrete Fourier Transform (DFT) and $\odot$ denotes element-wise multiplication.
The element with the maximal value in the original response map, obtained by inverse DFT, corresponds to the predicted target location.

\section{Group Feature Selection for DCF}
\subsection{GFS-DCF}
In DCF-based visual object tracking, multi-channel features are extracted from a large search window, in which only a small region is of interest.
In such a case, multi-channel image features are usually redundant and may bring uncertainty in the prediction stage.
To address this issue, spatial feature selection or regularisation has been widely used in existing DCF-based trackers, such as the use of fixed spatial masks~\cite{Galoogahi2017Learning,Lukezic2017Discriminative,li2018learning,Danelljan2016ECO}.
More recently, a learning-based adaptive mask~\cite{xu2018learning} has been proposed to inject spatial regularisation to DCF-based visual tracking, achieving the best performance on the VOT2018 public dataset~\cite{Kristan2018a}.
However, investigations aiming at reducing the information redundancy and noise across feature channels, especially as it applies to hundreds or thousands of deep CNN feature maps, are missing from the existing literature.
To close this gap, in this paper, we advocate a new feature selection method, namely Group Feature Selection (GFS), for DCF-based visual object tracking.

In contrast to previous studies, our GFS-DCF incorporates group feature selection, in both spatial and channel dimensions, in the original DCF optimisation task.
Additionally, a low-rank regularisation term is used to achieve temporal smoothness of the learned filters during the tracking process.
We assume that the learning of correlation filters is conducted for the $t$th frame and omit the subscript `$_t$' for simplicity.
The objective function of our GFS-DCF is:
\begin{equation}
\begin{aligned}\label{obj1}
\widetilde{\mathcal{W}} = &\arg\underset{\mathcal{W}}{\min}
\left\|\sum\limits_{k=1}^C\mathbf{W}^k\circledast{\mathbf{X}^k}-\mathbf{Y}\right\|_F^2 \\
& + \lambda_1\mathcal{R}_S(\mathcal{W})
+ \lambda_2\mathcal{R}_C(\mathcal{W}) + \lambda_3\mathcal{R}_T(\mathcal{W}),
\end{aligned}
\end{equation}
where $\mathcal{R}_S(\mathcal{W})$ is the spatial group regularisation term for spatial feature selection, $\mathcal{R}_C(\mathcal{W})$ is the group regularisation term for channel selection, $\mathcal{R}_T(\mathcal{W})$ is the temporal regularisation term and each $\lambda_i$ is a balancing parameter.
These regularisation terms are introduced in detail in the remaining part of this section and a solution of the above optimisation task is given in Section~\ref{sec_solution}.

\subsection{Group Spatial-Channel Regularisation}
Grouping is introduced into the model with the aim  of exploiting prior knowledge that is scientifically meaningful~\cite{huang2012selective}.
Considering the current nature of feature representations that are invariably multi-channel, and the spatial coherence of a tracked object, the grouping information is employed in $\mathcal{R}_{S}$ and $\mathcal{R}_{C}$ to achieve spatial-channel selection by allocating individual variables into specific groups with certain visual meaning (spatial location and channel attributes). This strategy has been demonstrated to be effective in visual data science~\cite{nie2010efficient,bach2012structured,yuan2006model,hu2017group,gui2017feature,wang2017deep,wang2018semi}.

To perform group feature selection for the spatial domain, we define the spatial regularisation term as:
\begin{equation}
\label{equ_ss}
\mathcal{R}_{S}(\mathcal{W})
=\sum\limits_{i=1}^{N}\sum\limits_{j=1}^{N}\left\|\mathbf{w}_{ij:}\right\|_2,
\end{equation}
in which we use $\ell_2$ norm to obtain the grouping attribute of each spatial location, calculated across all the feature channels.
To be more specific, we concatenate all the elements at the $i$th location of the first order and the $j$th location of the second order of the multi-channel feature tensor, $\mathcal{W}\in\mathbb{R}^{N \times N \times C}$, into a vector $\mathbf{w}_{ij:}=[w_{ij1},...,w_{ijC}]^\top$, as illustrated in Fig.~\ref{framework}. 
The grouping attribute is obtained by the $\ell_2$ norm and then the implicit $\ell_1$ norm of all the spatial grouping attributes is used to regularise the optimisation of correlation filters.
This naturally injects sparsity into the spatial domain by grouping all the elements across channels.
Such structured spatial sparsity enables robust group feature selection that reflects the joint contribution of features in the spatial domain.

In our preliminary experiments, we found that the proposed group feature selection in spatial domain is able to improve the performance of a DCF tracker when using hand-crafted features. 
However, the improvement is minor when we tried to impose spatial feature selection into deep CNN features.
We argue that the main reason is that an element in deep CNN feature maps stands for higher-level concepts thus performing spatial feature selection on such features cannot achieve fine-grained selection of the target region from the background.
For example we use the feature maps at the `res4x' layer of ResNet50~\cite{he2016deep} in the proposed method.
Each deep CNN feature map has the resolution of $13\times 13$, in which each pixel corresponds to a $16\times16$ region in the original input image.
To perform spatial selection on such a small resolution feature map cannot achieve very accurate spatial feature selection results.
But deep CNN features usually have many channels, with the cosequence of injecting information redundancy.
To address this issue, we propose channel selection by defining a group regularisation term in the channel dimension:
\begin{equation}
\label{equ_cs}
\mathcal{R}_{C}(\mathcal{W})
=\sum\limits_{k=1}^{C}\left\|\mathbf{W}^k\right\|_F,
\end{equation}
where we use the Frobenius norm to obtain the grouping attributes for feature channels $\{\mathbf{W}^k\}_{k=1}^{C}$. Note again that implicitly the constraint in (\ref{equ_cs}) is a sparsity inducing $\ell_1$ norm.
\begin{figure}[!t]
\centering
\subfloat[Hand-Crafted Features]
{\label{Impact_of_s}
\includegraphics[trim={80mm 20mm 88mm 21mm},clip,width=.495\linewidth]{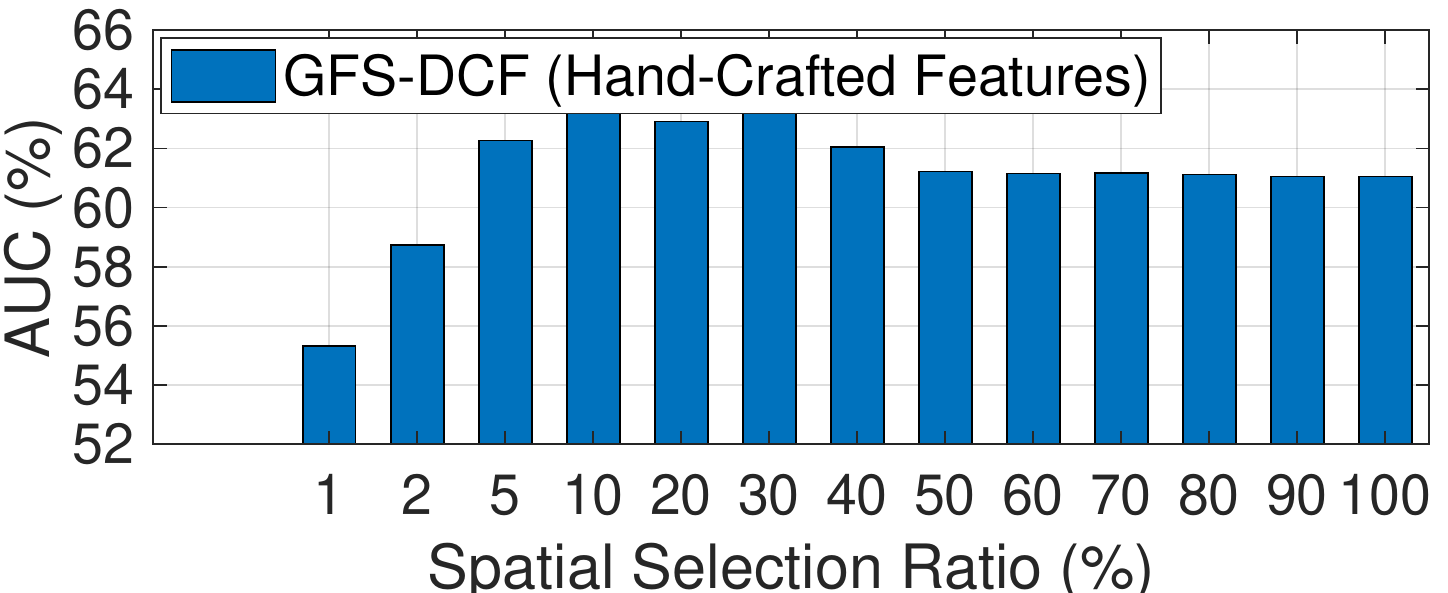}
\includegraphics[trim={80mm 20mm 88mm 21mm},clip,width=.495\linewidth]{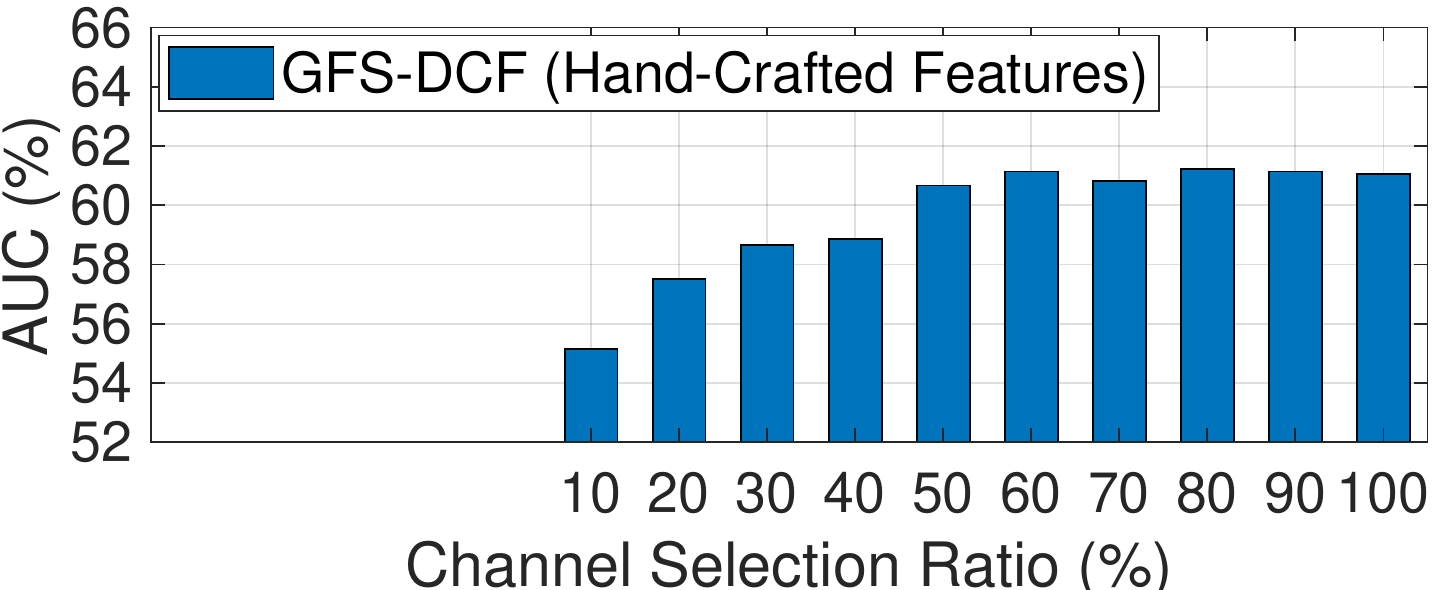}
}\\
\subfloat[Deep CNN Features]
{\label{Impact_of_c}
\includegraphics[trim={80mm 20mm 88mm 21mm},clip,width=.495\linewidth]{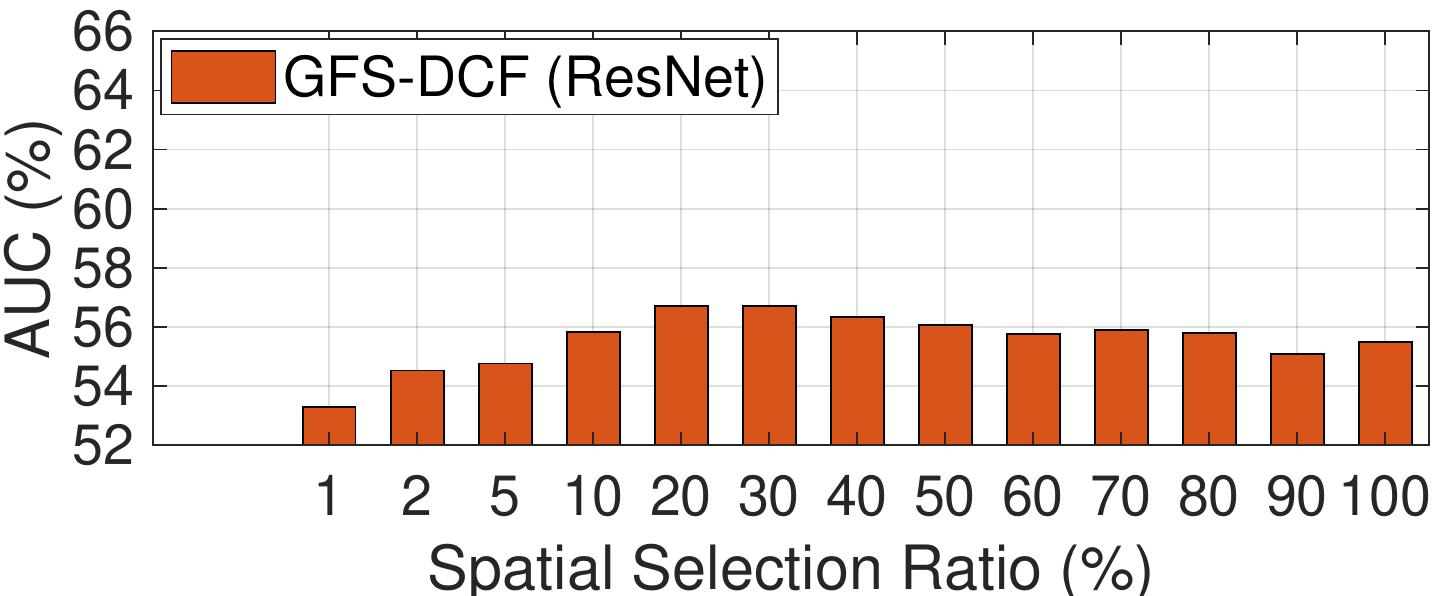}
\includegraphics[trim={80mm 20mm 88mm 21mm},clip,width=.495\linewidth]{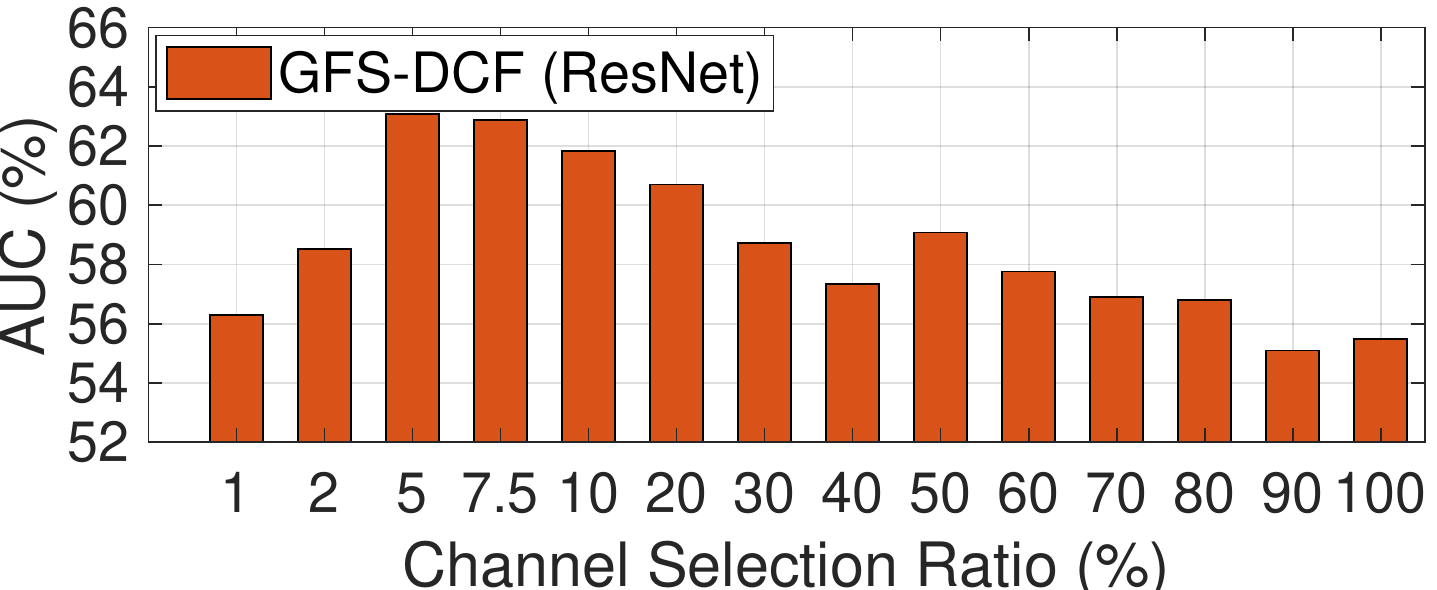}
}
\caption{A comparison of spatial and channel group feature selection on OTB2015 using either (a) hand-crafted or (b) deep CNN features, parameterised by selection ratio.}
\label{Impact_of_cs}
\end{figure}

In practice, to perform spatial/channel feature selection, we use the measures in Equ.~(\ref{equ_ss}) and Equ.~(\ref{equ_cs}).
Specifically, we first calculate the group attributes in spatial/channel domain and then eliminate the features across channel/spatial dimensions corresponding to a pre-defined proportion with the lowest grouping attributes.
This selection strategy has been commonly used in many previous studies~\cite{xu2011two,peng2015robust,song2018dictionary}.
Additionally, the proposed feature selection method is applied to each individual feature type separately.

To evaluate the effectiveness of the proposed spatial and channel group feature selection methods, we compare the proposed GFS-DCF with the classical DCF formulation on the OTB2015 dataset.
The results are shown in Fig.~\ref{Impact_of_cs}.
It should be noted that the bar at the selection ratio of 100\% stands for the original DCF tracker without feature selection, using either hand-crafted features or deep CNN features.
We use Colour Names, Intensity Channels and HOG for hand-crafted features, and ResNet50 for deep features.
Detailed experimental settings are introduced in Section~\ref{sec_setting}.
As reported in the figure, for hand-crafted features, impressive improvements are achieved with the spatial selection ratios ranging from $5\%\sim 40\%$.
But channel selection cannot improve the performance for hand-crafted features.
The only merit of performing channel selection on hand-crafted features is that we can maintain the performance when compressing the features to 60\% in size.
For deep features, the use of spatial feature selection only improves the performance marginally.
But, deep features benefit significantly from the channel selection regularisation, with the AUC increasing from $55.49\%$ to $63.07\%$ even when we only use 5\% of the original channels.
These results demonstrate that deep features are highly redundant across channels, and exhibit undesirable interference.
The evaluation validates the proposed spatial-channel group feature selection strategy.

As such, the proposed method offers a scope for dimensionality reduction by the proposed group spatial-channel regularisation, leading to performance boosting.
While hand-crafted features are extracted in a fixed manner with relatively high resolutions compared to deep features, different attributes are considered for different channels, with more redundancy and ambiguity in the spatial dimension.
The results support the conclusion that the tracking performance can be improved by using the proposed group-feature-selection-embedded filter learning scheme.

\subsection{Temporal Smoothness}
Despite the success of feature selection in many computer vision and pattern recognition tasks, it suffers from the instability of solutions, especially in the presence of information redundancy~\cite{meinshausen2010stability}. 
To mitigate this problem and take appearance variation into consideration~\cite{ross2008incremental}, we improve the robustness of learned correlation filters by injecting temporal smoothness. Specifically, a low-rank constraint is enforced on the estimates across video frames, so that the temporal coherence in the filter design is promoted.
We define the constraint as minimising:
\begin{equation}\label{lr}
rank\left(\mathbb{W}_{t}\right) - rank\left(\mathbb{W}_{t-1}\right),
\end{equation}
where $\mathbb{W}_{t} = [\textbf{vec}(\mathcal{W}_{1}),...,\textbf{vec}(\mathcal{W}_{t})] \in \mathbb{R}^{N^2C\times t}$ is a matrix, with each column storing the vectorised correlation filters, $\mathcal{W}$.

Here, the constraint~(\ref{lr}) imposes a low-rank property across frames because it impacts on the selection process since the second frame.
However, it is inefficient to calculate $rank\left(\mathbb{W}_{t}\right)$, especially in long-term videos with many frames. 
Therefore, we use its sufficient condition as a substitute:
\begin{equation}
d\left(\mathcal{W}_{t} - \mathcal{U}_{t-1}\right),
\end{equation}
where $\mathcal{U}_{t-1}=\sum_{k=1}^{t-1}\mathcal{W}_{k}/(t-1)$ is the mean over all the previous learned filters and $d$ is a distance metric.
The brief proof of sufficiency is provided as follows.

\noindent\textbf{Proof:} Given $\mathbb{W}_{t-1}$ and $\mathcal{U}_{t-1}$, the mean vector of $\mathbb{W}_{t}$ is influenced by $\mathcal{W}_{t}$. We denote $\check{\mathbf{w}}_{t}=\textbf{vec}\left( \sqrt{\frac{t-1}{t}}\left(\mathcal{W}_{t}-\mathcal{U}_{t-1}\right)\right)$. Expressing $\mathbb{W}_{t-1}$ in terms of its SVD as $\mathbb{W}_{t-1}=\textbf{U}_{t-1}\mathbf{\Sigma}_{t-1}\textbf{W}_{t-1}^\top$, we have,
\begin{equation}
\mathbb{W}_{t}=\left[
\begin{matrix}
\textbf{U}_{t-1},\textbf{vec}\left(\mathcal{W}_{t}\right)_{\bot}
\end{matrix}\right]\textbf{R}\left[
\begin{matrix}
\textbf{W}_{t-1}^\top&0\\0&1
\end{matrix}\right],
\end{equation}
and
\begin{equation}
\mathbf{R}=\left[
\begin{matrix}
\mathbf{\Sigma}_{t-1}&\mathbf{\Sigma}_{t-1}^\top\textbf{vec}\left(\mathcal{W}_{t}\right)\\
0&\check{\mathbf{w}}_{t}^{\top}\left(\textbf{I}-\textbf{U}_{t-1}\textbf{U}_{t-1}^{\top}\right)\check{\mathbf{w}}_{t}
\end{matrix}\right],
\end{equation}
where $\textbf{I}\in\mathbb{R}^{N^2Ct\times N^2Ct}$ is the identity matrix and $_{\bot}$ performs orthogonalisation of the vector, $\textbf{vec}\left(\mathcal{W}_{t}\right)$, to the matrix, $\mathbf{U}_{t-1}$. If $\mathcal{W}_{t} = \mathcal{U}_{t-1}$, then $\mathbf{\Sigma}_{t-1}$ dominates the eigenvalues of $\mathbf{R}$. Consequently, $\mathbf{R}$ shares the same eigenvalues as $\mathbb{W}_{t}$. $\Box$

Therefore, we propose to adaptively enforce the temporal low-rank property with the regularisation term:
\begin{equation}\label{r2}
\mathcal{R}_{\textrm{T}}\left(\mathcal{W}\right) =  \lambda_3\sum\limits_{k=1}^{C}\left\|\mathbf{W}^k_t-\mathbf{W}^k_{t-1}\right\|_F^2.
\end{equation}

\subsection{Solution}
\label{sec_solution}
Due to the convexity of the proposed formulation, we apply the augmented Lagrange method~\cite{Lin2010The} to optimise Equ.~(\ref{obj1}). 
Concretely, we introduce slack variable $\mathcal{W}^\prime=\mathcal{W}$ and construct the following Lagrange function:
\begin{equation}
\begin{aligned}
\label{lagrange}
\mathcal{L} =& \left\|\sum\limits_{k=1}^C\mathbf{W}^k_t\circledast{\mathbf{X}^k_t}-\mathbf{Y}\right\|_F^2 + \lambda_1\sum\limits_{k=1}^{C}\left\|\mathbf{W}^{\prime k}_t\right\|_F \\
+&\lambda_2\sum\limits_{i=1}^{N}\sum\limits_{j=1}^{N}\left\|\mathbf{w}^{\prime }_{ij_t}\right\|_2+\lambda_3\sum\limits_{k=1}^{C}\left\|\mathbf{W}^k_t-\mathbf{W}^k_{t-1}\right\|_F^2 \\
+& \frac{\mu}{2}\sum\limits_{k=1}^{C}\left\|\mathbf{W}^k_t-\mathbf{W}^{\prime k}_t+\frac{\mathbf{\Gamma}^k}{\mu}\right\|_F,
\end{aligned}
\end{equation}
where $\Gamma$ is the Lagrange multiplier sharing the same size as $\mathcal{X}$, $\mathbf{\Gamma}^k$ is its $k$-th channel, and $\mu$ is the corresponding penalty. 
Then the Alternating Direction Method of Multipliers~\cite{Boyd2011Distributed} is employed to perform iterative optimisation with guaranteed convergence as follows~\cite{petersen2008matrix}:
\begin{subequations}
\label{iter}
\begin{align}
\hat{\mathbf{w}}_{ij_t} &= \left(\mathbf{I}-\frac{\hat{\mathbf{x}}_{ij_t}\hat{\mathbf{x}}_{ij_t}^H}{\left(\lambda_3+\mu/2\right)N^2+\hat{\mathbf{x}}_{ij_t}^H\hat{\mathbf{x}}_{ij_t}}\right)\mathbf{q}\label{11},\\
w^{\prime k}_{ij_t} &= \max\left(0,1-\frac{\lambda_1}{\mu\left\|\mathbf{P}^k\right\|_F}
-\frac{\lambda_2}{\mu\left\|\mathbf{p}_{ij}\right\|_2}\right)p^k_{ij}\label{12},\\
\Gamma\ \ \  &= \Gamma + \mu\left(\mathcal{W}_t-\mathcal{W}_t^\prime\right)\label{13},
\end{align}
\end{subequations}
where $ \mathbf{q} = (\hat{\mathbf{x}}_{ij_t}\hat{y}_{ij}/N^2+\mu\hat{\mathbf{w}}^{\prime}_{ij_t}-\mu\hat{\mathbf{\gamma}}_{ij} + \lambda_3\hat{\mathbf{w}}_{ij_{t-1}}) / (\lambda_3+\mu)$
and
$p_{ij}^k=w_{ij}^k+\gamma_{ij}^k/\mu$.

\section{Evaluation}
\label{experiment}
\subsection{Implementation and Evaluation Settings}
\label{sec_setting}
We implement our GFS-DCF using MATLAB 2018a.
The speed of GFS-DCF is 8 frames per second (fps) on a platform with one Intel Xeon E5-2637 v3 CPU and NVIDIA GeForce GTX TITAN X GPU. 
We set $\lambda_1=10$ and $\lambda_2=1$ for group feature selection. In order to guarantee a fixed number of the selected channels and spatial units, we set up the channel selection ratio $r_c$ and spatial selection ratio $r_s$ to truncate the remaining channels and spatial units.
We extract hand-crafted features using Colour Names (CN), HOG, Intensity Channels (IC), and deep CNN features using ResNet-50~\cite{he2016deep,vedaldi2015matconvnet}.
For hand-crafted features, we set the parameters as $r_c=90\%$, $r_s=10\%$, $\lambda_3=16$ and $\alpha=0.6$.
For deep features, we set the parameters as $r_c=7.5\%$, $r_s=90\%$, $\lambda_3=12$ and $\alpha=0.05$.

We evaluated the proposed method on several well-known benchmarks, including OTB2013/OTB2015~\cite{Wu2013Online,Wu2015Object},  VOT2017/VOT2018~\cite{Kristan2017a,Kristan2018a} and TrackingNet Test dataset~\cite{muller2018trackingnet}, and compared it with a number of state-of-the-art trackers, such as VITAL~\cite{song2018vital}, MetaT~\cite{park2018meta}, ECO~\cite{Danelljan2016ECO}, MCPF~\cite{zhang2017multi}, CREST~\cite{song-iccv17-CREST}, BACF~\cite{Galoogahi2017Learning}, CFNet~\cite{valmadre2017end}, CACF~\cite{mueller2017context}, ACFN~\cite{choi2017attentional}, CSRDCF~\cite{Lukezic2017Discriminative}, C-COT~\cite{Danelljan2016Beyond}, Staple~\cite{Bertinetto2016Staple}, SiamFC~\cite{bertinetto2016fully},  SRDCF~\cite{Danelljan2015Learning}, KCF~\cite{Henriques2015High}, SAMF~\cite{li2014scale},  DSST~\cite{danelljan2017discriminative} and other advanced trackers in VOT challenges, \textit{i.e.},  CFCF~\cite{gundogdu2018good},  CFWCR~\cite{he2017correlation},  LSART~\cite{sun2018learning},  UPDT~\cite{bhat2018unveiling},  SiamRPN~\cite{zhu2018distractor},  MFT~\cite{Kristan2018a} and LADCF~\cite{xu2018learning}.

To measure the tracking performance, we follow the corresponding protocols~\cite{Wu2015Object,Kristan2016The,kristan2016novel}.
We use precision plot and success plot~\cite{Wu2013Online} for OTB2013 and OTB2015. 
Four numerical values, \textit{i.e.} centre location error (CLE), distance precision (DP), overlap precision (OP) and area under curve (AUC), are further employed to measure the performance.
For VOT2017 and VOT2018, we employ expected average overlap (EAO), accuracy value and robustness to evaluate the performance~\cite{Kristan2016The}.
For TrackingNet, we adopt success score, precision score and normalised precision to analyse the results~\cite{muller2018trackingnet}.

\begin{figure}[t]
\begin{center}
\includegraphics[trim={5mm 0mm 5mm 0mm},clip,width=0.495\linewidth]{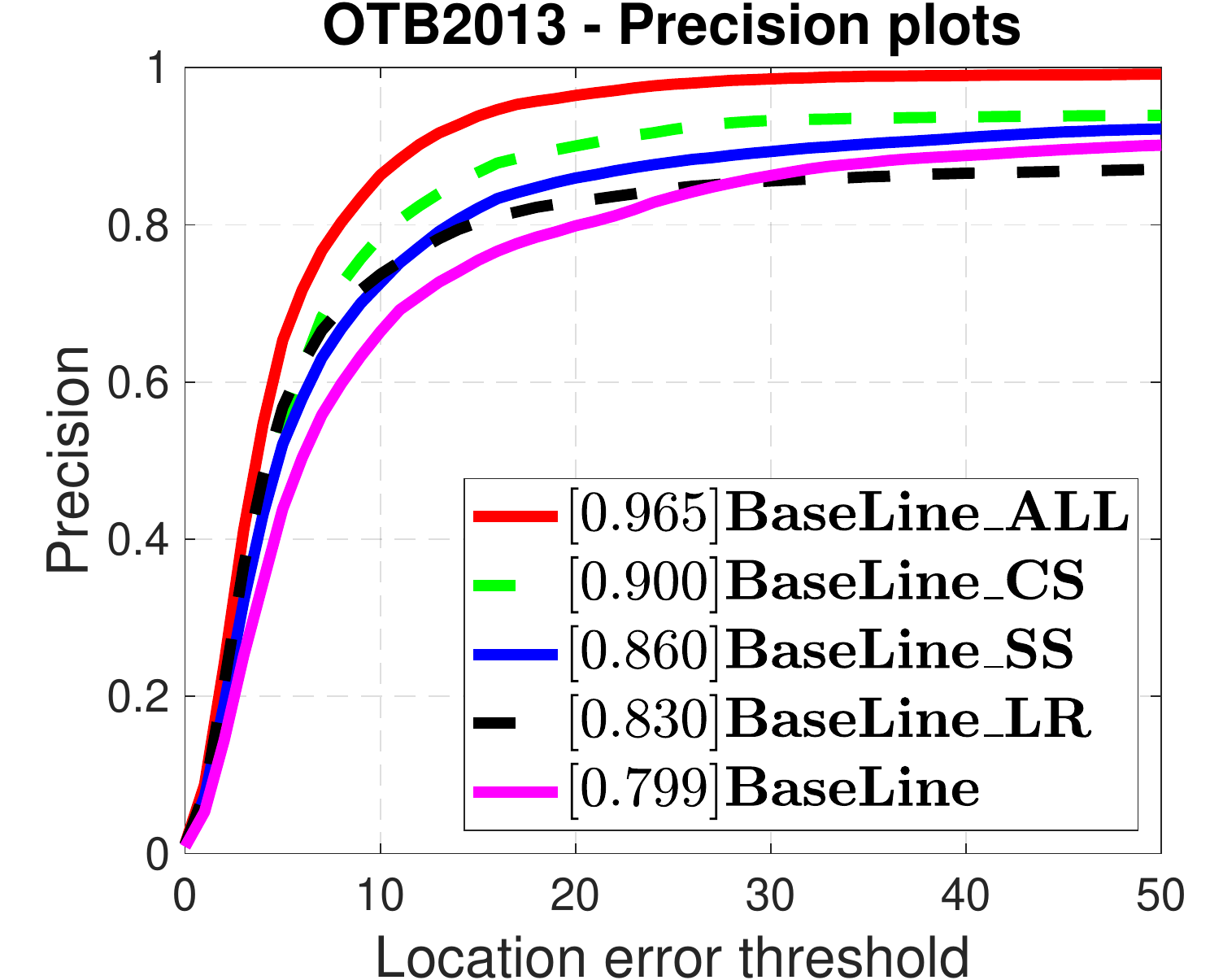}
\includegraphics[trim={5mm 0mm 5mm 0mm},clip,width=0.495\linewidth]{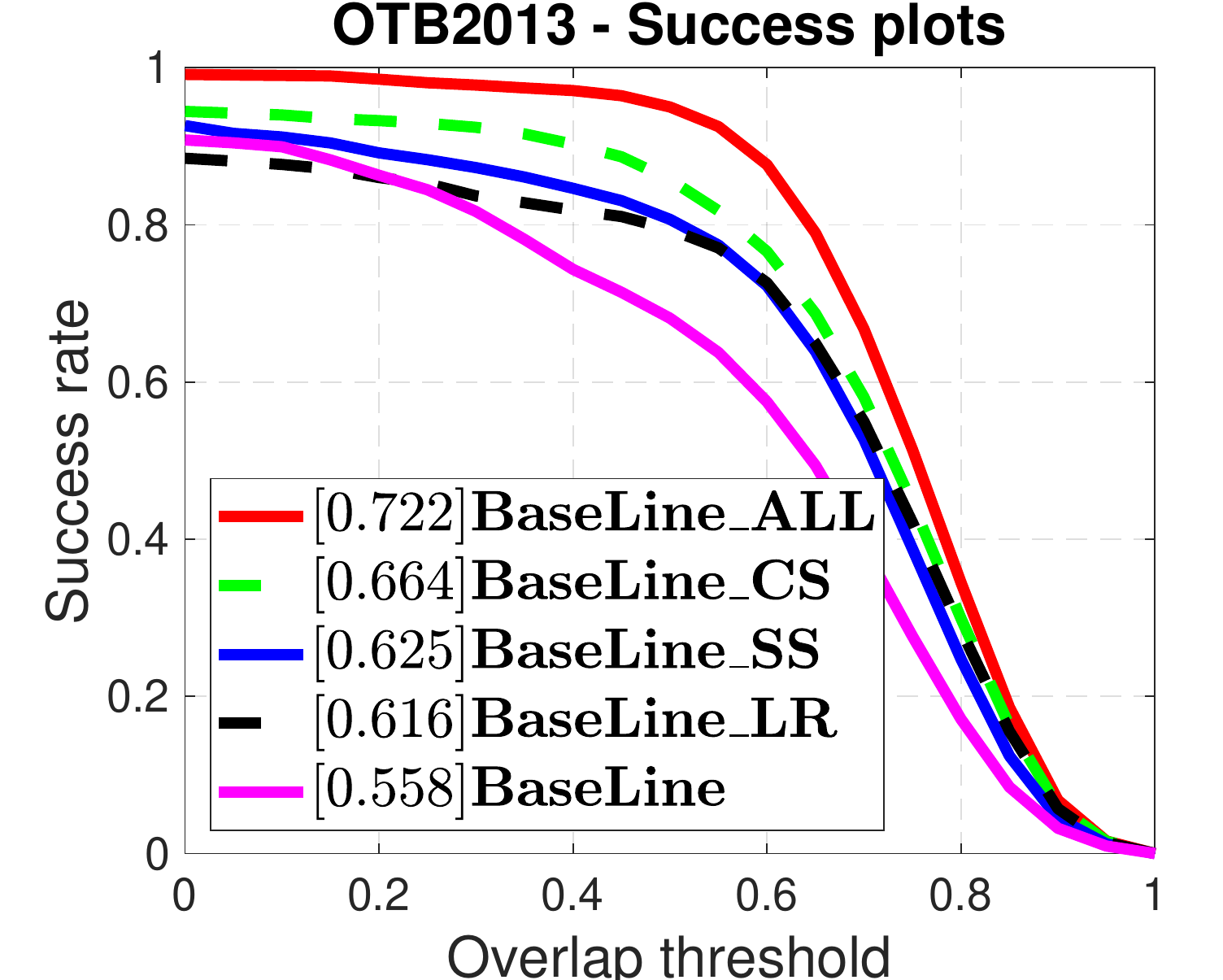}
\end{center}
\caption{A comparison of different regularisation terms in GFS-DCF, evaluated on OTB2013. The precision plots with \textbf{DP} and the success plots with \textbf{AUC} in the legends are presented.}
\label{component}
\end{figure}
\begin{table}[t]
\footnotesize
\renewcommand{\arraystretch}{1.3}
\caption{A comparison of different methods on different videos of OTB2015, in terms of the rank of the matrix formed by stacking all the vectorised filters of all the frames in a video. (The best three results are highlighted by {\color{red}{red}}, {\color{blue}{blue}} and {\color{brown}{brown}}.)}
\label{rank}
\centering
\begin{tabular}{l|cccc|c}
\hline
Video [\#frames] &KCF & CACF & ECO & C-COT & \textbf{GFS-DCF}\\
\hline
\hline
\textit{Deer} [71] & 71&14&{\color{brown}{\textbf{4}}}&{\color{blue}{\textbf{3}}}&{\color{red}{\textbf{2}}}\\
\textit{Basketball} [725]&526&134&{\color{brown}{\textbf{23}}}&{\color{blue}{\textbf{10}}}&{\color{red}{\textbf{9}}}\\
\textit{Boy} [602]&274&63&{\color{brown}{\textbf{19}}}&{\color{blue}{\textbf{8}}}&{\color{red}{\textbf{4}}}\\
\textit{David3} [252]&252&53&{\color{brown}{\textbf{8}}}&{\color{red}{\textbf{3}}}&{\color{blue}{\textbf{6}}}\\
\textit{Girl} [500]&267&57&{\color{brown}{\textbf{18}}}&{\color{blue}{\textbf{8}}}&{\color{red}{\textbf{5}}}\\
\textit{Suv} [945]&701&49&{\color{brown}{\textbf{16}}}&{\color{red}{\textbf{4}}}&{\color{blue}{\textbf{6}}}\\
\textit{Skater} [160]&160&38&{\color{brown}{\textbf{19}}}&{\color{red}{\textbf{3}}}&{\color{blue}{\textbf{5}}}\\
\textit{Woman} [597]&384&111&{\color{brown}{\textbf{15}}}&{\color{red}{\textbf{6}}}&{\color{blue}{\textbf{7}}}\\
\hline
\end{tabular}
\end{table}

\subsection{Ablation Study}
We first evaluate the effect of each innovative component in GFS-DCF, including the spatial selection term $\mathcal{R}_S$ (SS), channel selection term  $\mathcal{R}_C$ (CS) and low-rank temporal smooth term $\mathcal{R}_T$ (LR). 
The baseline is the original DCF tracker equipped with the same features (both hand-crafted and deep features) and updating rate as our GFS-DCF.
We construct 5 trackers,~\textit{i.e.}, BaseLine, BaseLine\_SS, BaseLine\_CS, BaseLine\_LR and BaseLine\_ALL, to analyse internal effectiveness.
The results evaluated on OTB2013 are reported in Fig.~\ref{component}.

According to the figure, the proposed channel selection, spatial selection and low-rank smoothness terms improve the performance of the classical DCF (BaseLine).
Compared with the classical DCF, the grouping channel/spatial selection terms $\mathcal{R}_C$/$\mathcal{R}_S$ (BaseLine\_CS/BaseLine\_SS) significantly improve the performance in terms of DP and AUC by $10.1\%$/$6.1\%$ and $10.6\%$/$6.7\%$.
The results are consistent with Fig.~\ref{Impact_of_cs}, demonstrating the redundancy and noise in the multi-channel representations and the advantage of performing group feature selection to achieve parsimony.
\begin{figure}[t]
\begin{center}
\includegraphics[trim={0mm 70mm 65mm 0mm},clip,width=1\linewidth]{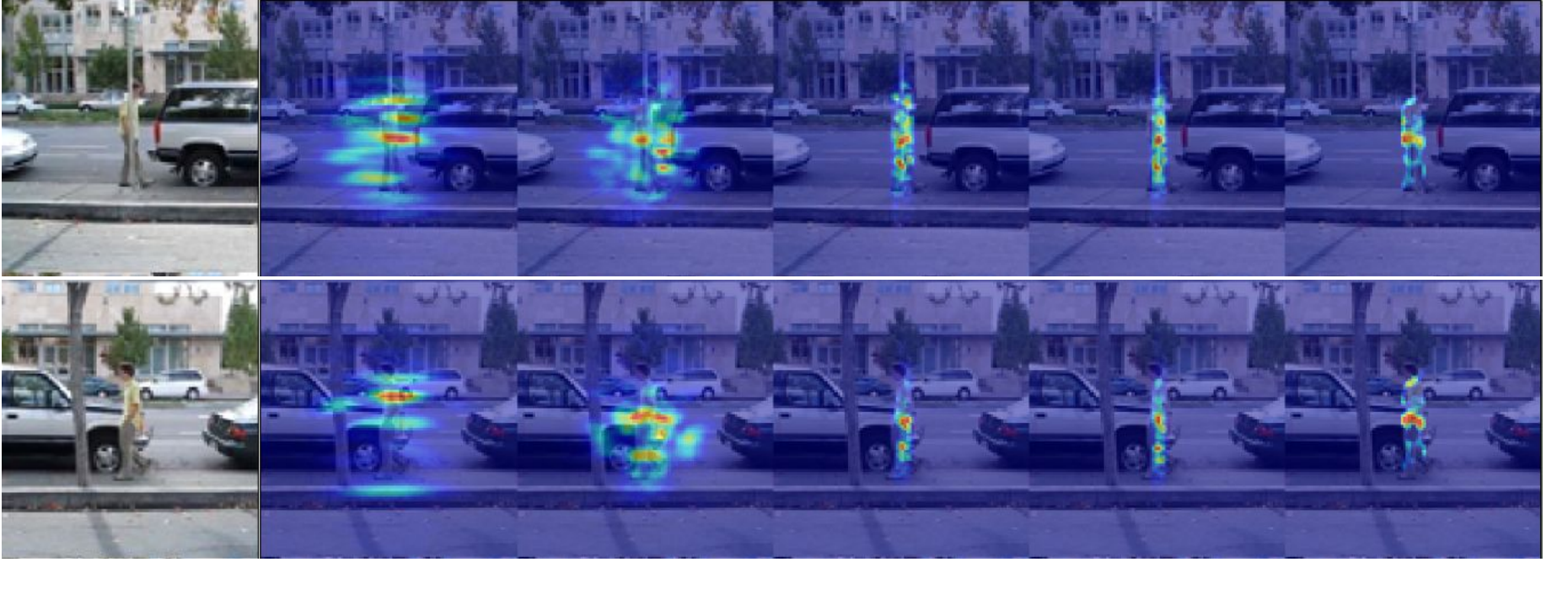}
\end{center}
\vspace{-0.4cm}
\footnotesize{\textbf{\ \ \ \ \ \ Input \ \ \ \ \ \ \ \ \ KCF\ \ \ \ \ \ \ \ \ \ CACF\ \ \ \ \ \ \ \ \  ECO\ \ \ \ \ \ \ \ \ C-COT \ \ \ \  \  GFS-DCF}}
\caption{Visualisation of filters using \textit{David3} in OTB2015. We visualise the corresponding filters in frame $\#50$ (the $1$st row) and $\#200$ (the $2$nd row). 
To better visualise the sparsity, we present the heat-maps of the obtained filters by gathering the energy across all the channels.}
\label{sparse}
\end{figure}
On the other hand, the low-rank temporal smoothness term $\mathcal{R}_T$ (BaseLine\_LR) also leads to improvement in the tracking performance.
Intuitively explained, a low-rank constraint across temporal frames enables the learned filters to become more invariant to appearance variations.
To verify the practical low-rank property, we further collect the filters of each frame, concatenate them together, and calculate the rank. 
To guarantee the quality of the collected filters, we only consider some simple sequences where all the involved trackers can successfully track the target across the entire frames, \textit{i.e.} the filters are effective in distinguishing the target from surroundings.
The results are presented in Table~\ref{rank}, which show that our simplified regularisation term, Equ.~(\ref{r2}), can achieve the low-rank property by only considering the filter model. 
Note, C-COT and ECO also share the low-rank property, achieved by taking into account historical appearance in the learning stage, but at the expense of increased complexity and storage.
\begin{figure}[!t]
\begin{center}
\includegraphics[trim={8mm 126mm 8mm 128mm},clip,width=1\linewidth]{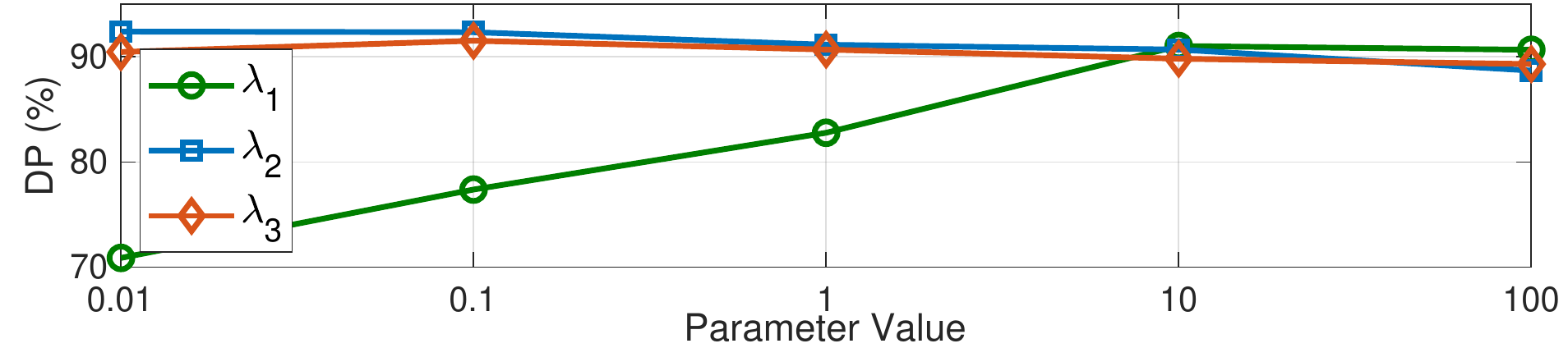}
\end{center}
\vspace{-0.4cm}
\caption{Impact of $\lambda_1$, $\lambda_2$ and $\lambda_3$, evaluated on OTB2015.}
\label{lambda}
\end{figure}

We further visualise the filters of 5 different trackers in Fig.~\ref{sparse}. Note that ECO and C-COT achieve sparsity by spatial regularisation, with more energy concentrating in the centre region. In contrast, our GFS-DCF realises sparsity without a pre-defined mask or weighting. The filters are adaptively shrunk to specific groups (channels/spatial units). Therefore, our GFS-DCF may shrink the elements even within the centre region.

In addition, we perform the sensitivity analysis of $\lambda_1$, $\lambda_2$ and $\lambda_3$. 
As shown in Fig.~\ref{lambda}, our GFS-DCF achieves stable performance with $\lambda_1$, $\lambda_2 \in [0.01, 100]$ and $\lambda_3\in [10, 100]$.
Though we have to set 7 parameters, the selection ratios are most essential, as shown in Fig.~\ref{Impact_of_cs}.
We employ threshold pruning operators to fix the ratio of selected spatial units and channels, enabling robustness against regularisation parameters.

Finally, the combination of all the components (BaseLine\_ALL) becomes our GFS-DCF tracker (Fig.~\ref{component}), which achieves the best performance compared with individual components.
The results demonstrate the effectiveness of the proposed grouping and low-rank formulations.

\subsection{Comparison with the State-of-the-art}
\begin{figure}[t]
\begin{center}
 \includegraphics[trim={5mm 0mm 5mm 0mm},clip,width=0.495\linewidth]{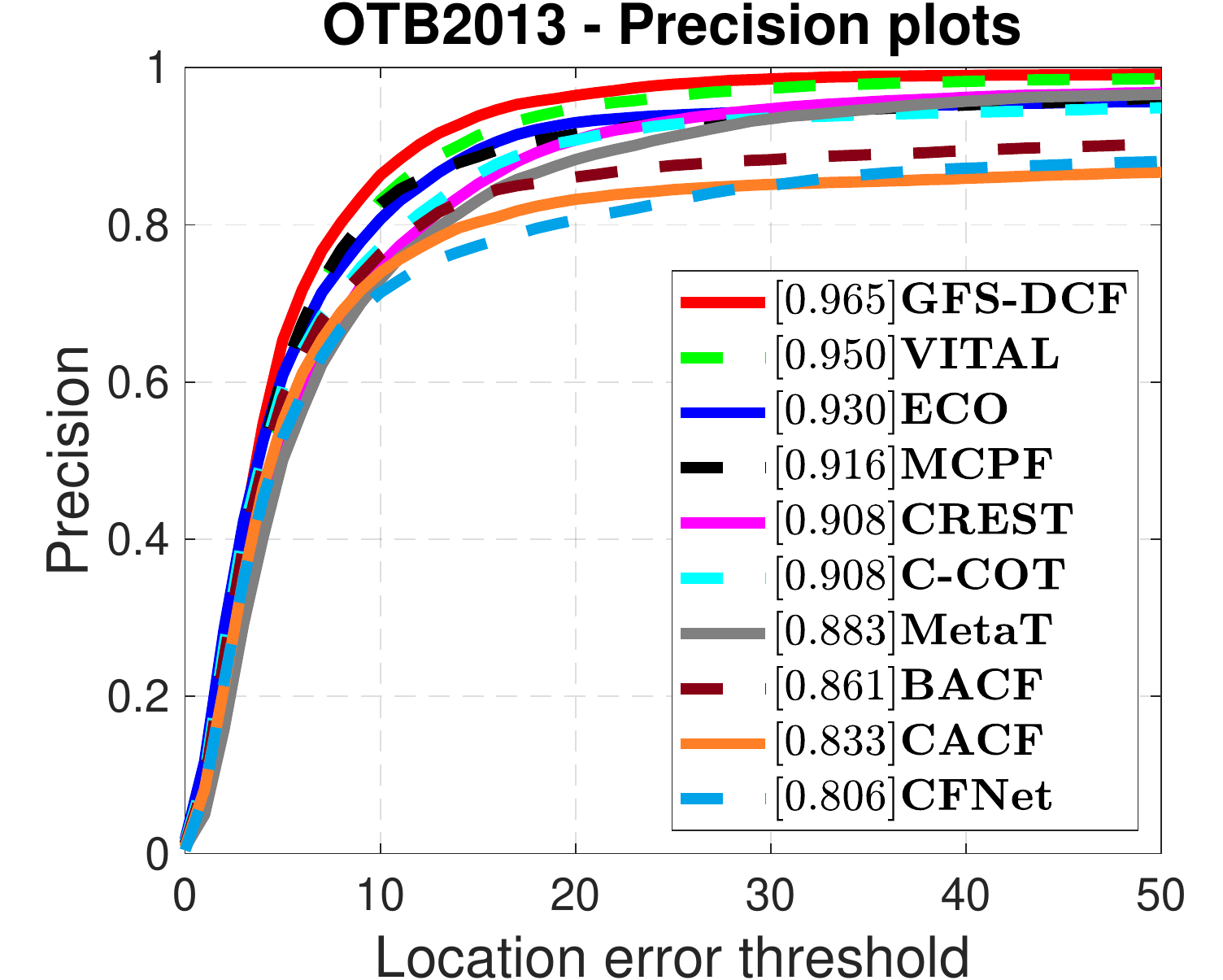}
 \includegraphics[trim={5mm 0mm 5mm 0mm},clip,width=0.495\linewidth]{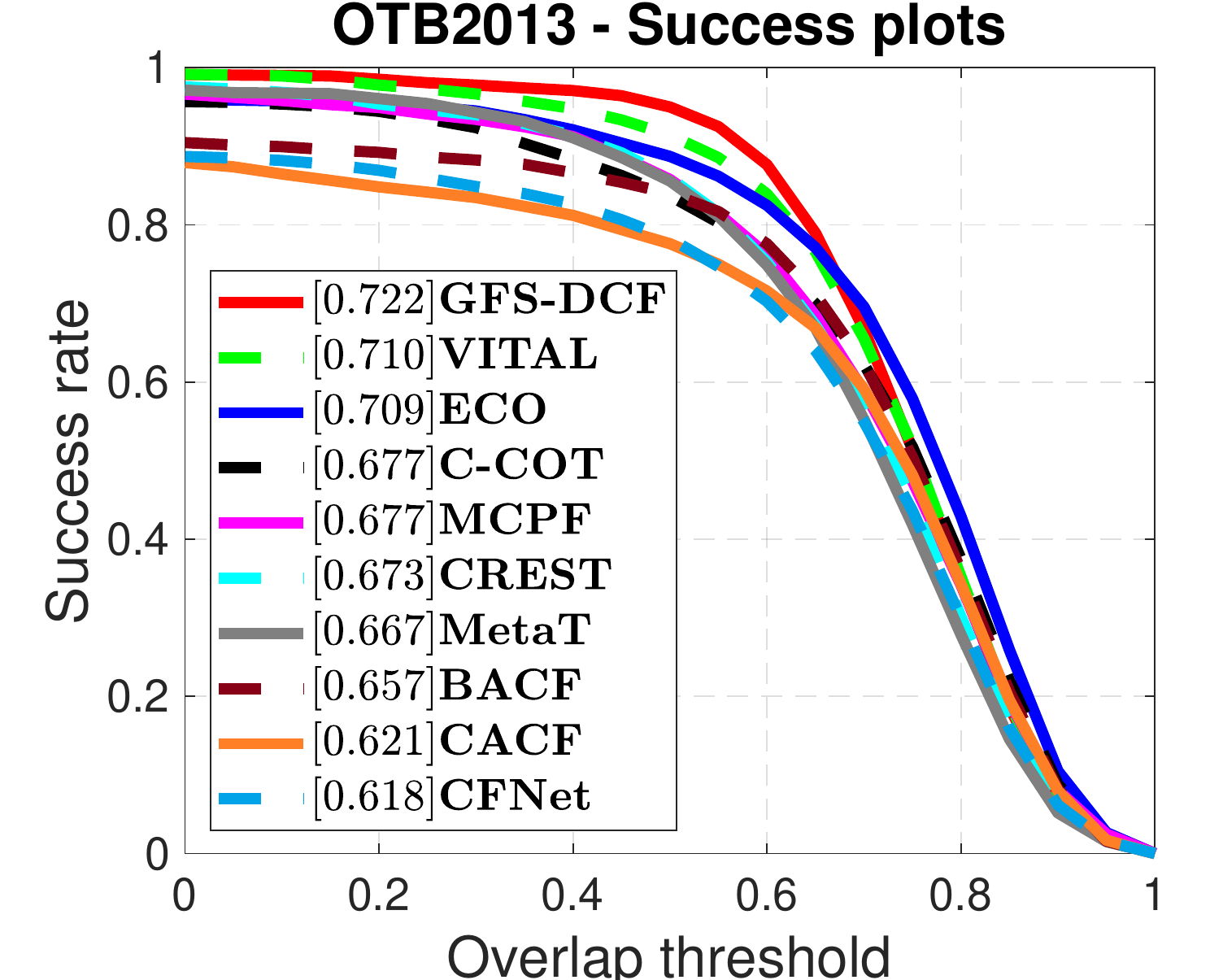}
 \\
 \includegraphics[trim={5mm 0mm 5mm 0mm},clip,width=0.495\linewidth]{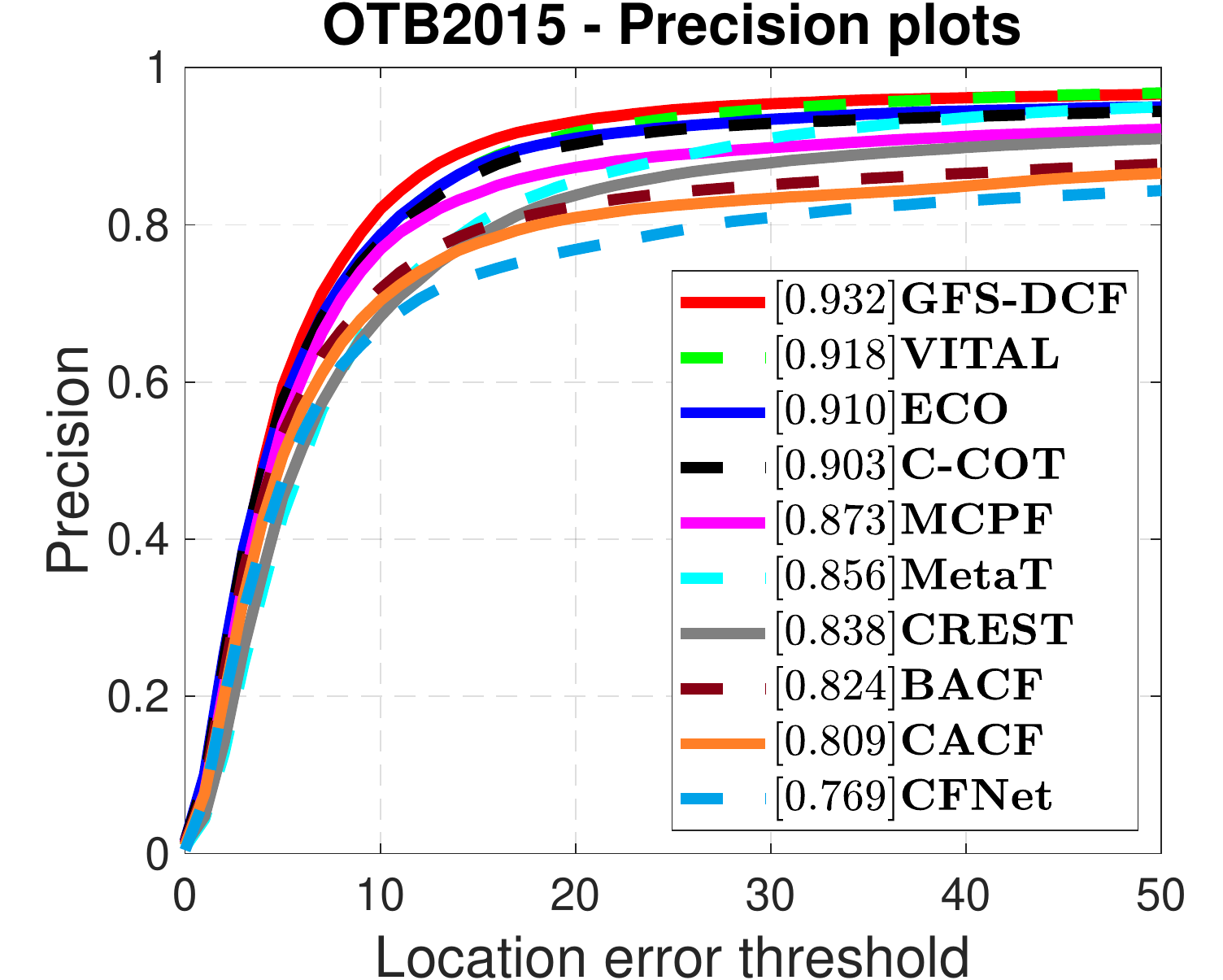}
 \includegraphics[trim={5mm 0mm 5mm 0mm},clip,width=0.495\linewidth]{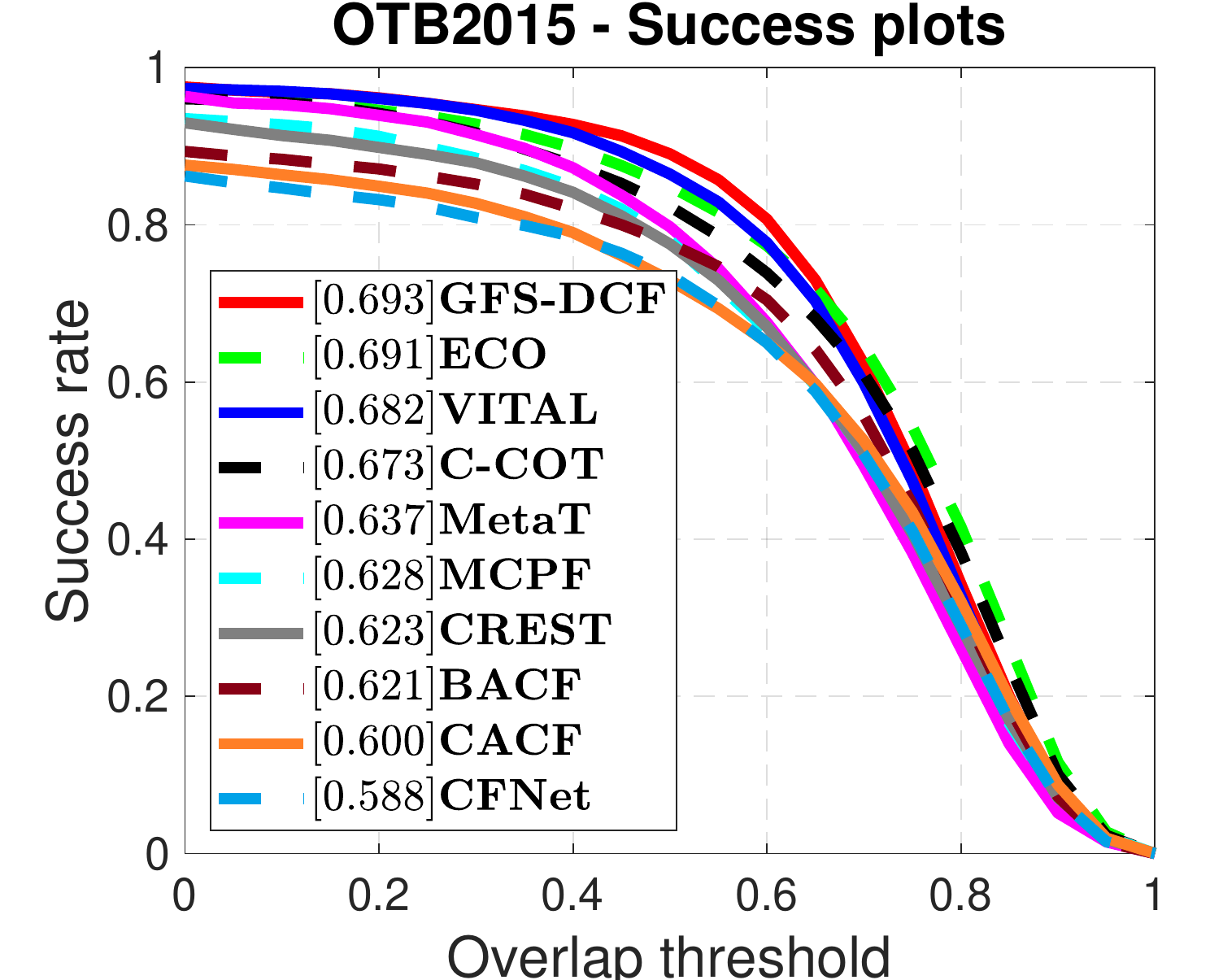}
 \\
\end{center}
   \caption{The experimental results on OTB2013, and OTB2015. The precision plots with \textbf{DP} reported in the figure legend (\textit{first column}) and the success plots with \textbf{AUC} reported in the figure legend (\textit{second column}) are presented. }\label{otb100}
\end{figure}

\begin{table}[t]
\footnotesize
\renewcommand{\arraystretch}{1.2}
\caption{Tracking results with different features on OTB2015.}
\label{detailed_results}
\centering
\begin{tabular}{l|l|ccc}
\hline
Feature & Method & OP & DP \\
\hline
\hline
\multirow{5}{*}{HOG} & BACF & 77.6\% & 82.4\% \\
& CSRDCF & 70.5\% & 79.4\%\\
& SRDCF & 71.1\% & 76.7\% \\
& LADCF & \textbf{78.5\%} & 83.1\% \\
& GFS-DCF & 78.2\% & \textbf{85.2\%} \\
\hline
\multirow{3}{*}{HOG+CN}& ECO & 78.0\% & 85.1\% \\
& C-COT & 75.7\% & 84.1\%\\
& GFS-DCF & \textbf{81.5\%} & \textbf{86.3\%} \\
\hline
\multirow{3}{*}{HOG+CN+VGG-M}& ECO & 84.9\% & 91.0\%\\
& C-COT & 82.3\% & 90.3\%\\
& GFS-DCF & \textbf{85.5\%} & \textbf{91.2}\%\\
\hline
\end{tabular}
\end{table}

\begin{table*}[!t]
\footnotesize
\renewcommand{\arraystretch}{1.3}
\caption{The \textbf{OP} and \textbf{CLE} results on OTB2013, TB50 and OTB2015. (The best three results are highlighted by {\color{red}{red}}, {\color{blue}{blue}} and {\color{brown}{brown}}.)}
\label{otb}
\centering
\begin{tabular}{c|l|ccccccccc}
\hline
 & & KCF & SAMF & DSST & SRDCF & SiamFC & Staple & C-COT & CSRDCF & ACFN\\
\hline
\hline
\multirow{ 3}{*}{$\begin{matrix}\textrm{ \textbf{OP/CLE}}\\($\%$/\textrm{pixels})\end{matrix}$} &\textbf{OTB2013} &  60.8/36.3 & 69.6/29.0 & 59.7/39.2 & 76.0/36.8 & 77.9/29.7 & 73.8/31.4 & 83.7/15.6 & 74.4/31.9 & 75.0/18.7\\
&\textbf{TB50}    & 47.7/54.3 & 59.3/40.5 & 45.9/59.5 & 66.1/42.7 & 68.0/36.8 & 66.5/32.3 & 80.9/{\color{red}{\textbf{12.3}}} & 66.4/30.3 & 63.2/32.1\\
&\textbf{OTB2015}    & 54.4/45.1 & 64.6/34.6 & 53.0/49.1 & 71.1/39.7 & 73.0/33.2 & 70.2/31.8 & 82.3/{\color{brown}{\textbf{14.0}}} & 70.5/31.1 & 69.2/25.3\\
\textbf{SPEED} (fps) & & {\color{red}{\textbf{82.7}}} & 11.5 & 15.6 & 2.7 & 12.6 & {\color{blue}{\textbf{23.8}}} & 2.2 & 4.6 & 13.8\\
\hline
& & CACF & CFNet & BACF & CREST & MCPF & ECO & MetaT & VITAL & \textbf{GFS-DCF}\\
\hline
\hline
\multirow{ 3}{*}{$\begin{matrix}\textrm{  \textbf{OP/CLE}}\\($\%$/\textrm{pixels})\end{matrix}$}  &\textbf{OTB2013} & 77.6/29.8 & 78.3/35.2 & 84.0/26.2 & 86.0/{\color{brown}{\textbf{10.2}}} & 85.8/11.2 & {\color{brown}{\textbf{88.7}}}/16.2 & 85.6/11.5 & {\color{blue}{\textbf{91.4}}}/{\color{blue}{\textbf{7.4}}} & {\color{red}{\textbf{95.0}}}/{\color{red}{\textbf{5.92}}} \\
&\textbf{TB50}    & 68.1/36.3 & 68.8/36.7 & 70.9/30.3 & 68.8/32.6 & 69.9/30.9 & {\color{brown}{\textbf{81.0}}}/13.2 & 73.7/17.0 & {\color{blue}{\textbf{81.3}}}/{\color{brown}{\textbf{12.5}}} & {\color{red}{\textbf{82.8}}}/{\color{blue}{\textbf{12.4}}}\\
&\textbf{OTB2015}   & 73.0/33.1 & 73.6/36.0 & 77.6/28.2 & 77.6/21.2 & 78.0/20.9 & {\color{brown}{\textbf{84.9}}}/14.8 & 79.8/14.2 & {\color{blue}{\textbf{86.5}}}/{\color{red}{\textbf{9.9}}} & {\color{red}{\textbf{89.0}}}/{\color{blue}{\textbf{10.3}}}\\
\textbf{SPEED} (fps) & & {\color{brown}{\textbf{18.1}}} & 8.7 & 16.3 & 10.1 & 0.5 & 12.5 & 0.8 & 1.3 & 7.8\\
\hline
\end{tabular}
\end{table*}

\begin{table*}[t]
\footnotesize
\renewcommand{\arraystretch}{1.3}
\caption{Tracking results on VOT2017/VOT2018. (The best three results are highlighted by {\color{red}{red}}, {\color{blue}{blue}} and {\color{brown}{brown}}.)}
\label{vot17}
\centering
\begin{tabular}{l|cccccccc|c}
\hline
& ECO~\cite{Danelljan2016ECO} & CFCF~\cite{gundogdu2018good} & CFWCR~\cite{he2017correlation} & LSART~\cite{sun2018learning} & UPDT~\cite{bhat2018unveiling} & SiamRPN~\cite{zhu2018distractor} & MFT~\cite{Kristan2018a} & LADCF~\cite{xu2018learning} & GFS-DCF\\
\hline
\hline
\textbf{EAO} & 0.280 & 0.286 & 0.303 & 0.323  & 0.378 & 0.383 & {\color{brown}{\textbf{0.385}}} & {\color{blue}{\textbf{0.389}}} & {\color{red}{\textbf{0.397}}} \\
\textbf{Accuracy} & 0.483 & 0.509 & 0.484 & 0.493 & {\color{blue}{\textbf{0.536}}} & {\color{red}{\textbf{0.586}}} & 0.505 & 0.503 & {\color{brown}{\textbf{0.511}}} \\
\textbf{Robustness} & 0.276 & 0.281 & 0.267 & 0.218 & 0.184 & 0.276 & {\color{red}{\textbf{0.140}}} & {\color{brown}{\textbf{0.159}}} & {\color{blue}{\textbf{0.143}}} \\
\hline 
\end{tabular}
\end{table*}
\noindent \textbf{OTB}
We report the precision and success plots for OTB2013 and OTB2015 in Fig.~\ref{otb100}. 
Overall, our GFS-DCF outperforms all the other state-of-the-art trackers in terms of DP and AUC.
Compared with the second best tracker, GFS-DCF achieves the improvements by $1.5\%/1.2\%$ and $1.4\%/0.2\%$ (in DP/AUC) on OTB2013 and OTB2015, respectively.

To achieve a fair comparison of mathematical formulations, we also compared our method with the state-of-the-art trackers using the same features on OTB2015. As shown in Table~\ref{detailed_results}, our GFS-DCF performs better than almost all the other approaches regardless of the features used, demonstrating the advantage of the proposed method.

We also present the detailed OP, CLE and speed (fps) of all the involved trackers on OTB2013, TB50 and OTB2015 in Table~\ref{otb}. 
On OTB2013, our GFS-DCF tracker achieves the OP of $95.0\%$ and CLE of $5.92\ pixels$.
Compared with the recent VITAL and MetaT trackers based on end-to-end deep neural networks, our performance gain is $3.6\%/1.48\ pixel$ and $8.4\%/5.58\ pixels$ in terms of OP and CLE, respectively.
On TB50, GFS-DCF performs better than C-COT (by $1.9\%$) in terms of OP but with a lower CLE (by $0.1\ pixel$).
In addition, on OTB2015, our tracker outperforms many recent trackers,~\textit{i.e.} CSRDCF (by $18.5\%/20.8\ pixels$), CACF (by $16.0\%/22.8\ pixels$), C-COT (by $6.7\%/3.7\ pixels$), BACF (by $11.4\%/17.9\ pixels$) and ECO (by $4.1\%/4.5\ pixels$) in terms of OP/CLE. 

\noindent \textbf{VOT}
Table~\ref{vot17} presents the results obtained on VOT2017/VOT2018 dataset~\cite{Kristan2018a}. Our method achieves the best EAO score, 0.397, outperforming recent advanced trackers,~\textit{e.g.}, LADCF, UPDT and SiamRPN.
In addition, the reported Accuracy (0.511) and Robustness (0.143) results of GFS-DCF are also within the top three, demonstrating the effectiveness of the proposed group selection framework.

\noindent \textbf{TrackingNet}
We also report the results generated by the TrackingNet~\cite{muller2018trackingnet} evaluation server (511 test sequences) in Table~\ref{tracking_net_results}. Our GFS-DCF achieves $71.97\%$ in normalised precision, demonstrating its advantages as compared with the other state-of-the-art methods.

\begin{table}[!t]
\footnotesize
\renewcommand{\arraystretch}{1.3}
\caption{Evaluation on the TrackingNet test set.}
\label{tracking_net_results}
\centering
\begin{tabular}{l|ccc}
\hline
Method & Success & Precision & Normalised Precision\\
\hline\hline
CACF~\cite{mueller2017context} & 53.59\% & 46.72\% & 60.84\%\\
ECO~\cite{Danelljan2016ECO}  & 56.13\% & 48.86\% & 62.14\% \\
MDNet~\cite{nam2016learning} & \textbf{61.35}\% & 55.53\% & 71.00\%\\
\textbf{GFS-DCF} & 60.90\% & \textbf{56.57\%} & \textbf{71.79\%}\\
\hline
\end{tabular}
\end{table}

In conclusion, the proposed GFS-DCF tracking method achieves advanced performance, as compared to the state-of-the-art trackers, with favourable speed.


\section{Conclusion}\label{conclusion}
We proposed an effective appearance model with outstanding performance by learning spatial-channel group-sparse discriminative correlation filters, constrained by low-rank approximation across successive frames.
By reformulating the appearance learning model so as to incorporate group-sparse regularisation and a temporal smoothness constraint, we achieved adaptive temporal-spatial-channel filter learning on a low dimensional manifold with enhanced interpretability of the learned model. 
The extensive experimental results on visual object tracking benchmarks demonstrate the effectiveness and robustness of our method, compared with the state-of-the-art trackers. 
The diversity of hand-crafted and deep features in terms of spatial and channel dimensions is examined to support the conclusion that different selection strategies should be performed on different feature categories.

\noindent \textbf{Acknowledgment}
This work was supported in part by the National Natural Science Foundation of China (61672265, U1836218, 61876072), the 111 Project of Ministry of Education of China (B12018), the EPSRC Programme Grant (FACER2VM) EP/N007743/1, EPSRC/dstl/MURI project EP/R018456/1 and the NVIDIA GPU Grant Program.

{\small
\bibliographystyle{ieee}
\bibliography{bibabbre}
}

\end{document}